\documentclass[letterpaper]{article} 
\usepackage{aaai2026}  
\usepackage{times}  
\usepackage{helvet}  
\usepackage{courier}  
\usepackage[hyphens]{url}  
\usepackage{graphicx} 
\usepackage{natbib}  
\usepackage{caption} 
\usepackage{algorithm}
\usepackage{algorithmic}
\usepackage{color} 

\usepackage{booktabs}
\usepackage{multirow}
\usepackage{amsmath, amssymb}
\usepackage{bm}

\newcommand{\textud}[1]{\underline{#1}}
\newcommand{\tsb}[1]{\textsubscript{{#1}}}

\newcommand{\tdr}{\textsuperscript{\textdagger}}
\newcommand{\ddr}{\textsuperscript{\textdaggerdbl}}
\newcommand{\str}{\textsuperscript{*}}

\newcommand{\pms}[1]{\textsubscript{\(\pm\){#1}}}
\newcommand{\use}{\(\surd\)}
\newcommand{\notuse}{{\Large\(\times\)}}

\newcommand{\ie}{\textit{i.e.}}

\newcommand{\acc}{Acc.\(\uparrow\)}
\newcommand{\forg}{Forgetting\(\downarrow\)}

\newcommand{\eqname}{Eq.}
\usepackage{newfloat}
\usepackage{listings}
\DeclareCaptionStyle{ruled}{labelfont=normalfont,labelsep=colon,strut=off} 
\floatstyle{ruled}
\newfloat{listing}{tb}{lst}{}
\floatname{listing}{Listing}
\title{Attention Retention for Continual Learning with Vision Transformers}
\author{
	Yue Lu\textsuperscript{\rm 1}\textsuperscript{\textdagger},
	Xiangyu Zhou\textsuperscript{\rm 1}\textsuperscript{\textdagger},
	Shizhou Zhang\textsuperscript{\rm 1}\footnotemark[1],
	Yinghui Xing\textsuperscript{\rm 1},
	Guoqiang Liang\textsuperscript{\rm 1},
	Wencong Zhang\textsuperscript{\rm 2}\thanks{Shizhou Zhang and Wencong Zhang are co-corresponding authors. ~\newline~\textdagger\;Yue Lu and Xiangyu Zhou are co-first authors.}
}
\affiliations{
	\textsuperscript{\rm 1}School of Computer Science, Northwestern Polytechnical University, China\\
	\textsuperscript{\rm 2}Hikrobot Co., Ltd., China\\
	zgxd@mail.nwpu.edu.cn, 2021300119@mail.nwpu.edu.cn, szzhang@nwpu.edu.cn, xyh\_7491@nwpu.edu.cn, gqliang@nwpu.edu.cn, zhangwencong@hikrobotics.com
}

\begin{document}

\maketitle

\begin{abstract}
Continual learning (CL) empowers AI systems to progressively acquire knowledge from non-stationary data streams. However, \emph{catastrophic forgetting} remains a critical challenge.
In this work, we identify \emph{attention drift} in Vision Transformers as a primary source of catastrophic forgetting, where the attention to previously learned visual concepts shifts significantly after learning new tasks.
Inspired by neuroscientific insights into the selective attention in the human visual system, we propose a novel attention-retaining framework to mitigate forgetting in CL.
Our method constrains attention drift by explicitly modifying gradients during backpropagation through a two-step process:
1) extracting attention maps of the previous task using a layer-wise rollout mechanism and generating instance-adaptive binary masks,
and 2) when learning a new task, applying these masks to zero out gradients associated with previous attention regions, thereby preventing disruption of learned visual concepts.
For compatibility with modern optimizers, the gradient masking process is further enhanced by scaling parameter updates proportionally to maintain their relative magnitudes.
Experiments and visualizations demonstrate the effectiveness of our method in mitigating catastrophic forgetting and preserving visual concepts.
It achieves state-of-the-art performance and exhibits robust generalizability across diverse CL scenarios.
\end{abstract}

\section{Introduction}
Humans are capable of continuously accumulating and integrating knowledge in dynamic environments.
Similarly, for AI systems, the ability to learn continuously over time is essential for achieving long-term adaptation and intelligent evolution.
\emph{Continual learning} (CL), which aims to enable AI systems to learn adaptively in non-stationary data streams, akin to how humans acquire and develop knowledge.

\begin{figure}[!t]
	\centering
	\includegraphics[]{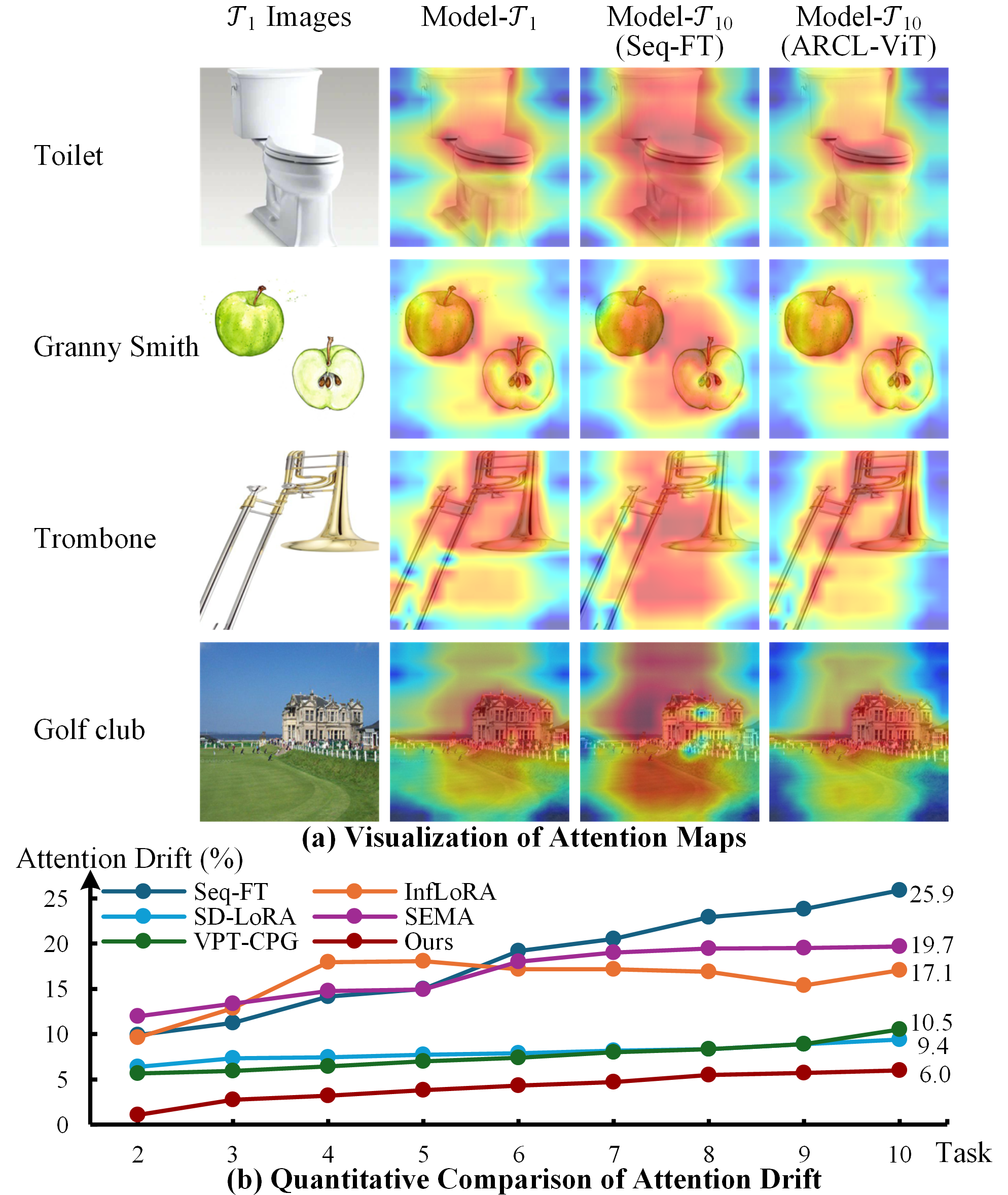}
	\caption{(a) Visualization of attention maps on the first task's (\(\mathcal{T}_1\)) samples. Columns two to four show attention maps from ViT models that are (i) only fine-tuned on \(\mathcal{T}_1\), (ii) sequentially fine-tuned over 10 tasks (Model-\(\mathcal{T}_{10}\) Seq-FT), and (iii) trained with our method over 10 tasks. Seq-FT leads to severe attention drift, while our method effectively preserves the original attention and thus mitigates forgetting.
		(b) Quantitative comparison of attention drift relative to \(\mathcal{T}_1\). 
	}
	\label{fig1}
\end{figure}

\emph{Catastrophic forgetting} remains the central challenge in continual learning for artificial neural networks.
Extensive efforts have been devoted to addressing this issue.
For instance, replay-based methods~\cite{ICaRLIncremental2017Rebuffi, MemoryReplay2018Wu} store and replay previously learned samples.
Regularization-based approaches~\cite{OvercomingCatastrophic2017Kirkpatrick, ContinualLearning2019Zeng} constrain changes in important parameters.
Expansion-based methods~\cite{DynamicallyExpandable2021Yan, CODAPromptCOntinual2023Smith} allocate independent network branches for new tasks.
These strategies mimic various aspects of human learning, such as memory consolidation, synaptic plasticity regulation, and the formation of new neural connections.
They offer valuable insights into building continual learning systems that resemble biological intelligence.

In the visual system, limited neural resources force the brain to selectively process the sensory inputs.
\cite{NeuralActivities2012Zhang} report that the primary visual cortex (also known as V1) generates a saliency map through neural activity to highlight attention-attracting regions in the visual field.
This selective attention remains consistently anchored on the discriminative features of visual concepts, even as new concepts are acquired.
Such inherent stability ensures that previously learned knowledge is preserved and not disrupted by subsequent learning, forming the biological foundation for non-forgetting in human visual cognition.
Inspired by this neuroscientific insight, we argue that \emph{preventing attention drift from previously learned concepts} is essential for CL systems to overcome catastrophic forgetting.
However, Vision Transformers~\cite{ImageWorth2021Dosovitskiy} (ViTs) exhibit severe \emph{attention drift} after learning new tasks, as illustrated in the third column of \figurename~\ref{fig1}(a).
To address this issue and enhance the alignment between ViT attention mechanism and biological perception, we propose ARCL-ViT, an \textud{A}ttention-\textud{R}etaining method that explicitly constrains attention drift on previous tasks during the learning of new tasks, thereby alleviating catastrophic forgetting.

However, directly optimizing the objective of preventing attention drift requires storing previous task data and is computationally infeasible.
Instead, we approximately achieve this objective by \emph{suppressing gradients contributing to shifting previous attention} during backpropagation, which eliminates the need for old task data.
This is achieved through two steps:
1) Attention Mask Generation. After completing a task, we extract the attention maps corresponding to the learned categories and generate binary masks that identify their critical attention regions.
In this step, we introduce a layer-wise rollout mechanism for attention map extraction and develop an instance-adaptive thresholding strategy to dynamically identify salient attention regions.
2) Gradient Masking. When learning the next task, we use the masks to zero out the gradients associated with these regions, so as to prevent parameter updates from disrupting the model’s original attention to previous visual concepts.
To ensure compatibility with modern optimizers, we further scale the parameter updates proportionally to maintain the relative magnitude between the masked and original gradients.
As a result, our method suppresses attention drift and mitigates forgetting, as shown in the last column of \figurename~\ref{fig1}(a).

Overall, we summarize our contributions as follows:
\begin{itemize}
	\item We identify attention drift in Vision Transformers as a primary source of catastrophic forgetting in continual learning, and reveal the necessity of preserving selective attention to previous visual concepts.
	\item We propose a novel framework for retaining attention, which constrains attention drift via gradient masking and employs a layer-wise rollout and adaptive thresholding strategy to generate effective attention masks.
	\item Our method is extensively evaluated on challenging continual learning benchmarks, achieving state-of-the-art performance and robust generalization across diverse pre-training schemes and long-sequence settings.
\end{itemize}


\section{Related Work}

\textbf{Replay-based methods} mitigate catastrophic forgetting by emulating the hippocampal mechanism of memory reactivation, which involves periodically revisiting prior knowledge to consolidate learning.
These methods can be broadly categorized into two types.
1) Memory-based replay retains and reuses samples from previous tasks, with the key challenge being the efficient selection of exemplars.
For instance, \cite{ICaRLIncremental2017Rebuffi} use a herding strategy to select samples whose features closely approximate those of the entire training set.
2) Generative replay~\cite{MemoryReplay2018Wu, IncrementalLearning2019Xiang}, which synthesizes samples from past tasks using generative models to avoid storing real data.
However, replay-based methods often raise concerns regarding data privacy and storage overhead.
Consequently, recent CL approaches utilizing pre-trained ViTs have predominantly focused on the replay-free setting.

\textbf{Expansion-based methods} create new branches when learning new tasks, mimicking the formation of new synaptic connections in the brain.
In CNN-based approaches, such as DER~\cite{DynamicallyExpandable2021Yan}, expansion is implemented by adding new convolutional branches while freezing previously learned feature extractors.
In ViT-based approaches, the expansion mechanism is implemented through dynamic extension of prompt pools~\cite{LearningPrompt2022Wang}.
Representative methods such as DualPrompt~\cite{DualPromptComplementary2022Wang} and NNPrompt~\cite{NearestNeighborClass2026Lu} learn a task-specific prompt for each task, which is then stored in a prompt pool.
During inference, the model selects a relevant prompt from the pool for the input instance and incorporates the prompt into the prediction process.
Therefore, each prompt essentially serves as a parameterized branch tailored to a specific task.

\textbf{Regularization-based methods}, inspired by biological synaptic plasticity that enables the brain to adapt to new environments, facilitate continual learning by adjusting parameter plasticity.
Representative approaches include EWC~\cite{OvercomingCatastrophic2017Kirkpatrick} and SI~\cite{ContinualLearning2017Zenke}, which estimate the importance of each parameter to previous tasks and penalize changes to critical weights when learning new tasks.
Another line of regularization-based methods focuses on orthogonal projections, such as OWM~\cite{ContinualLearning2019Zeng}, NSCL~\cite{TrainingNetworks2021Wang, lu2024visual, BalancingStability2022Kong} and GPM~\cite{GradientProjection2021Saha, PromptGradient2024Qiao}.
These methods aim to project model parameters onto a subspace that preserves the outputs from previous tasks while learning new ones, with differences in how they implement orthogonality.

Our approach can be categorized as a regularization-based method, as it directly constrains gradients during learning new tasks.
By masking gradients associated with prior attention regions, it reduces the disruption of previously acquired attention and effectively mitigates forgetting.

\begin{figure*}[t]
	\centering
	\includegraphics[]{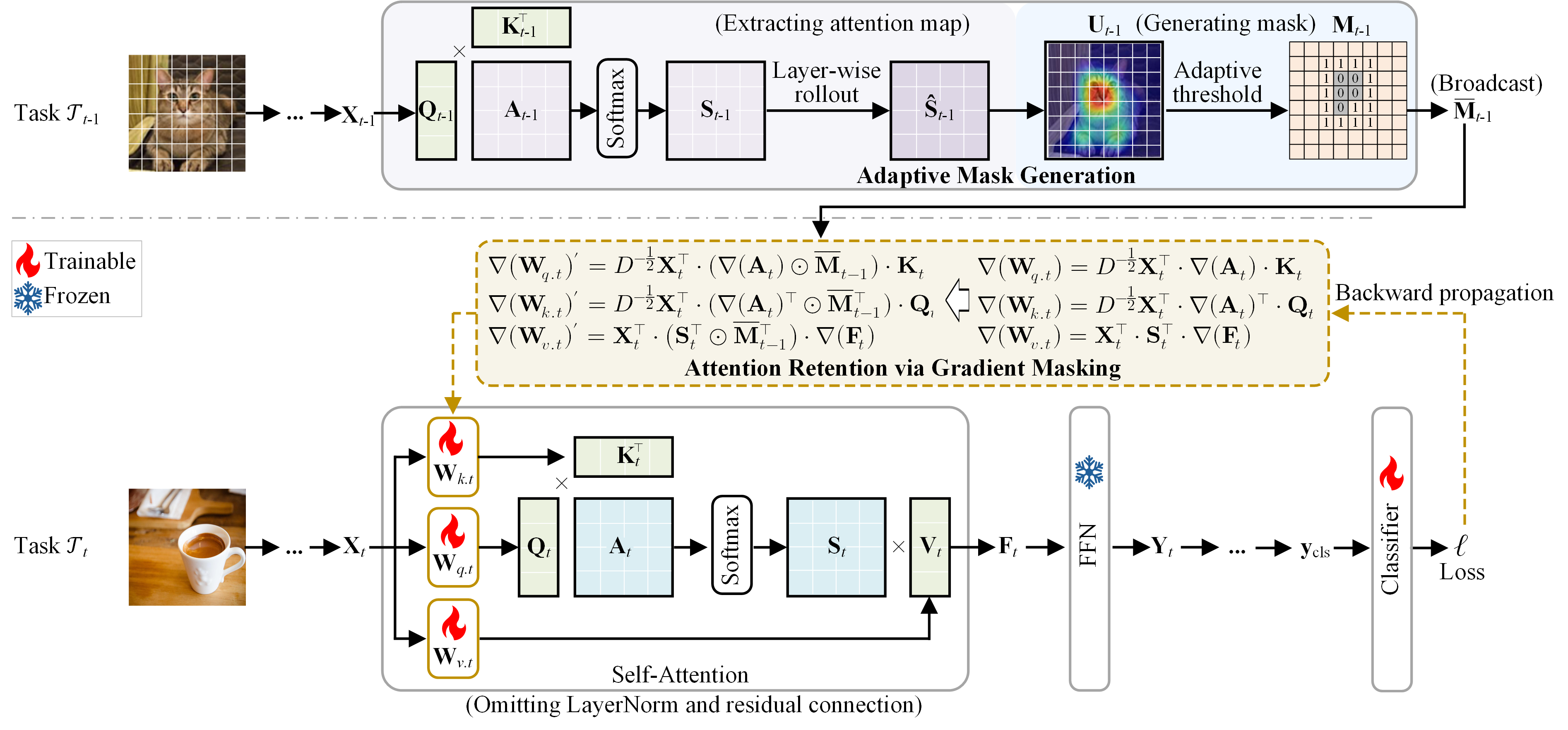}
	\caption{Illustration of our proposed attention-retaining continual learning (ARCL-ViT) framework. After finishing \( \mathcal{T}_{t-1} \), the model extracts attention maps \( \mathbf{U}_{t-1} \) and generates the mask \( \bar{\mathbf{M}}_{t-1} \) to identify attention regions. During subsequent training in \( \mathcal{T}_{t} \), the mask \( \bar{\mathbf{M}}_{t-1} \) is used to selectively zero out gradients in the corresponding attention regions. The masked gradients \( \nabla(\mathbf{W}_{q.t})^{\prime}, \nabla(\mathbf{W}_{k.t})^{\prime}, \nabla(\mathbf{W}_{v.t})^{\prime} \) are used to update the learnable weights \( \mathbf{W}_{q.t}, \mathbf{W}_{k.t}, \mathbf{W}_{v.t} \).}
	\label{fig2}
\end{figure*}

\section{Preliminaries}
\subsection{Continual Learning Problem Formulation}
Under the continual learning protocol, a model is trained sequentially on a series of \(T\) tasks \(\mathcal{T}_1, \mathcal{T}_2, \ldots, \mathcal{T}_T\).
For the \(t\)-th task \(\mathcal{T}_t\), the training set is denoted as \(\mathcal{D}_t = \{ (\mathbf{x}_t^{(n)}, y_t^{(n)}) \}_{n=1}^{|\mathcal{T}_t|}\), where \(\mathbf{x}_t^{(n)}\) is the \(n\)-th input sample with label \(y_t^{(n)}\), and \(|\mathcal{T}_t|\) denotes the number of instances in this task.
We focus on the class-incremental learning scenario, where the label space \(\mathcal{Y}_t\) of each task \(\mathcal{T}_t\) is disjoint from those of other tasks, \textit{i.e.}, \(\mathcal{Y}_t \cap \mathcal{Y}_{t'} = \emptyset\) for \(t \ne t'\).
After completing training on \(\mathcal{T}_t\), the corresponding training set \(\mathcal{D}_t\) becomes unavailable for future tasks.
The model is expected to accurately classify test samples from all previously learned classes, without access to task identity at inference time.

\subsection{Vision Transformer}
We adopt the Vision Transformer~\cite{ImageWorth2021Dosovitskiy} (ViT) as our base model.
A given image is first divided into \( N \) non-overlapping patches and linearly projected into \( D \)-dimensional tokens.
Then a class token is prepended to the sequence, followed by the addition of positional embeddings.
Next, the resulting sequence is fed into \( L \) Transformer blocks.
Within the \( l \)-th block \( f^l(\cdot) \), the input tokens first undergo a LayerNorm, and the normalized tokens \( \mathbf{X}^l \!\in\! \mathbb{R}^{(N+1) \times D} \) are transformed through three projections:
\begin{equation}
	\mathbf{Q}^l = \mathbf{X}^l \mathbf{W}_q^l, \quad
	\mathbf{K}^l = \mathbf{X}^l \mathbf{W}_k^l, \quad
	\mathbf{V}^l = \mathbf{X}^l \mathbf{W}_v^l,
	\label{eq:qkv}
\end{equation}
where \( \mathbf{W}_q^l, \mathbf{W}_k^l, \mathbf{W}_v^l \!\in\! \mathbb{R}^{D \times D} \).
The self-attention output is computed as:
\( \mathbf{F}^l \!=\! \mathrm{softmax} \left( \frac{\mathbf{Q}^l (\mathbf{K}^l)^\top}{\sqrt{D}} \right) \mathbf{V}^l\).
As shown in the ``Self-Attention'' module in \figurename~\ref{fig2}, this operation can be decomposed into three steps:
\begin{equation}
		\mathbf{A}^l = D^{-\frac{1}{2}} \mathbf{Q}^l (\mathbf{K}^l)^\top,
		\mathbf{S}^l = \mathrm{softmax}(\mathbf{A}^l),
		\mathbf{F}^l = \mathbf{S}^l \mathbf{V}^l \!.
	\label{eq:decomposed_attn}
\end{equation}
\( \mathbf{F}^l \) is then passed through another LayerNorm followed by a feedforward network (FFN) to produce the output sequence \( \mathbf{Y}^l \) for this block.
Finally, the class token from the last layer's output (denote as \( \mathbf{y}^L_{\mathrm{cls}} \!\in\! \mathbb{R}^D \)) is extracted and fed into the classifier \( g(\cdot) \) and loss function \( \ell(\cdot) \) for optimization.

In our setting, we employ a ViT model that has been pre-trained on large-scale datasets.
During continual learning, we \emph{freeze all pre-trained parameters except for the projection weights \( \mathbf{W}_q^l, \mathbf{W}_k^l, \mathbf{W}_v^l \) in Transformer block \( f^l(\cdot) \) and the classifier \( g(\cdot) \)}.
This design allows our model to adjust the attention weights to new tasks while preserving general representations from the pre-trained backbone.

\section{Method}
\subsection{Overview of Our Approach}
Our objective is to prevent attention drift from the previous task \( \mathcal{T}_{t-1} \) when optimizing parameters for the current task \( \mathcal{T}_t \).
Let \( \psi^l(\cdot) \) denote the function extracting the attention map at the $l$-th layer. This goal is formulated as:
\begin{equation}
	\begin{aligned}
		& \min_{\mathbf{W}_{\theta.t}^l} \mathbb{E}_{(\mathbf{x}_t,y_t) \sim \mathcal{D}_t} [\ell(g(f^l(\mathbf{x}_t; \mathbf{W}^l_{\theta.t}), y_t))], \\
		& \mathrm{s.t.}~ \| \psi^l(\mathbf{x}_{t-1};\! \mathbf{W}_{\theta.t}^l) - \psi^l(\mathbf{x}_{t-1};\! \mathbf{W}_{\theta.t-1}^l) \| = 0,
	\end{aligned}
\end{equation}
where \( \mathbf{x}_{t-1} \) denotes samples from \( \mathcal{D}_{t-1} \), and \( \theta \) represents \( \{q,\!k,\!v\} \).
Directly optimizing this objective is infeasible due to: 1) the complexity of \( \psi^l(\cdot) \) makes closed-form solutions intractable, and 2) the need to store \( \mathcal{D}_{t-1} \) incurring heavy storage overhead.
Instead, we analyze how gradients impact attention regions and observe that updates along attention gradient directions directly alter attention values.
By suppressing gradients in previously attended regions, we thus prevent such shifts and preserve prior attention patterns (see Appendix-``Theoretical Justification''), which enables us to approximately achieve the objective in a data-free manner.

As illustrated in Figure~\ref{fig2}, our approach consists of two steps: \emph{Adaptive Mask Generation} and \emph{Gradient Masking}.
The first step incorporates a layer-wise attention rollout and instance-adaptive thresholding to generate a mask \( \bar{\mathbf{M}}_{t-1} \) for previous task, which is then applied in the second step as a gradient mask to modify gradients.
In the following, we first assume a suitable attention mask \( \bar{\mathbf{M}}_{t-1} \) is already available and introduce the proposed gradient masking for attention retention, followed by the procedure for mask generation.

\subsection{Attention Retention via Gradient Masking}
We first analyze the relationship between the gradients \( \nabla(\cdot) \) \textit{w.r.t.} the parameters \( \mathbf{W}_{\theta.t} \) and the attention matrix \( \mathbf{A}_t \) for the current task \( \mathcal{T}_t \). For simplicity, the layer superscript \( l \) is omitted. 
Based on \eqname~\eqref{eq:decomposed_attn} which defines \(\mathbf{A}\), we have the following derivations:
\( \nabla(\mathbf{Q}_t) = D^{-\frac{1}{2}} \nabla(\mathbf{A}_t) \mathbf{K}_t \),
\( \nabla(\mathbf{K}_t) = D^{-\frac{1}{2}} \nabla(\mathbf{A}_t)^\top \mathbf{Q}_t \).
Similarly, from \eqname~\eqref{eq:decomposed_attn} for \(\mathbf{F}\), we obtain:
\( \nabla(\mathbf{V}_t) = \mathbf{S}_t^\top \nabla(\mathbf{F}_t) \).
Building on the QKV computation in \eqname~\eqref{eq:qkv}, we further derive the relationship between the gradients \( \nabla(\mathbf{W}_{q.t}), \nabla(\mathbf{W}_{k.t}), \nabla(\mathbf{W}_{v.t}) \) and \( \mathbf{A}_t \):
\begin{align}
	\left\{
	\begin{aligned}
			\nabla(\mathbf{W}_{q.t}) &= D^{-\frac{1}{2}} \mathbf{X}_t^{\top} \cdot \nabla(\mathbf{A}_t) \cdot \mathbf{K}_t,  \\
			\nabla(\mathbf{W}_{k.t}) &= D^{-\frac{1}{2}} \mathbf{X}_t^{\top} \cdot \nabla(\mathbf{A}_t)^\top \cdot \mathbf{Q}_t, \\
			\nabla(\mathbf{W}_{v.t}) &= \mathbf{X}_t^{\top} \cdot \mathbf{S}_t^{\top} \cdot \nabla(\mathbf{F}_t).
		\end{aligned}
	\right.
	\label{eq:grad_nomod}
\end{align}

From the above equations, we observe that the attention matrix \(\mathbf{A}_t\) (and its activated counterpart \(\mathbf{S}_t\)) directly influences the gradients \(\nabla(\mathbf{W}_{\theta.t})\). 
Under standard training, these gradients are applied via gradient descent to update the weights \( \mathbf{W}_{\theta.t} \). 
Consequently, the attention computed by the updated weights shifts toward the \emph{discriminative regions} (\ie, the most salient object features reflecting the visual concept, such as the head for a cat) of \(\mathcal{D}_t\), thereby improving accuracy on the current task.
However, the updated model may lose focus on \(\mathcal{D}_{t-1}\), resulting in catastrophic forgetting and degraded performance.

To address this issue, we \emph{remove gradients in the regions corresponding to the previous task’s attention during backpropagation when training the current task}.
Specifically, when completing the training of task \(\mathcal{T}_{t-1}\), we extract the attention map \(\mathbf{U}_{t-1}\) and construct a binary mask \( \bar{\mathbf{M}}_{t-1} \) by setting high-attention regions to 0 and low-attention regions to 1.
During training of \(\mathcal{T}_t\), we perform an element-wise multiplication (\( \odot \)) between \(\bar{\mathbf{M}}_{t-1}\) and \(\mathbf{A}_t\) (or \(\mathbf{S}_t\)), thereby eliminating gradients corresponding to the previous task’s attention regions in the updates of \( \mathbf{W}_{\{q.t,\,k.t,\,v.t\}} \).
Formally, the gradient masking is formulated as:
\begin{align}
	\left\{
	\begin{aligned}
		\nabla(\mathbf{W}_{q.t})^{\prime} &= D^{-\frac{1}{2}} \mathbf{X}_t^{\top} \cdot (\nabla(\mathbf{A}_t) \odot \bar{\mathbf{M}}_{t-1}) \cdot \mathbf{K}_t,  \\
		\nabla(\mathbf{W}_{k.t})^{\prime} &= D^{-\frac{1}{2}} \mathbf{X}_t^{\top} \cdot (\nabla(\mathbf{A}_t)^\top \odot \bar{\mathbf{M}}_{t-1}^\top) \cdot \mathbf{Q}_t, \\
		\nabla(\mathbf{W}_{v.t})^{\prime} &= \mathbf{X}_t^{\top} \cdot (\mathbf{S}_t^{\top} \odot \bar{\mathbf{M}}_{t-1}^\top) \cdot \nabla(\mathbf{F}_t),
	\end{aligned}
	\right.
	\label{eq:grad_mod}
\end{align}
where \( \nabla(\cdot)^{\prime} \) represents the masked gradients.

Nevertheless, the optimizer does not update parameters directly based on the raw gradients, \ie, \(\Delta \mathbf{W}_{\theta.t} \!\neq\! \nabla(\mathbf{W}_{\theta.t})\), where \(\Delta \mathbf{W}_{\theta.t}\) denotes the parameter update.
For example, the Adam optimizer computes \(\Delta \mathbf{W}_{\theta.t}\) based on the first and second moments of past gradients. This can lead to excessively large or small updates when using the masked gradients, potentially hindering optimization.
Therefore, we further adapt the proposed gradient masking to optimizers.

Specifically, we aim to scale the parameter updates so that the ratio between the masked and original updates matches the ratio of the corresponding gradients, formulated as:
\begin{equation}
	\frac{\Delta \mathbf{W}_{\theta.t}^{\prime}}{\Delta \mathbf{W}_{\theta.t}~} = \frac{\nabla(\mathbf{W}_{\theta.t})^{\prime}}{\nabla(\mathbf{W}_{\theta.t})~},
\end{equation}
where \(\Delta \mathbf{W}_{\theta.t}\) denotes the parameter update computed by the optimizer using the unmasked gradient \(\nabla(\mathbf{W}_{\theta.t})\), while \(\Delta \mathbf{W}_{\theta.t}^{\prime}\) represents the expected parameter update.
By rearranging the above equation, we derive:
\begin{equation}
	\Delta \mathbf{W}_{\theta.t}^{\prime} = \frac{\nabla(\mathbf{W}_{\theta.t})^{\prime}}{\nabla(\mathbf{W}_{\theta.t})~} \odot \Delta \mathbf{W}_{\theta.t}.
\end{equation}
We then use \(\Delta \mathbf{W}_{\theta.t}^{\prime}\) to update the parameters:
\begin{equation}
	\mathbf{W}_{\theta.t}^{<s>} = \mathbf{W}_{\theta.t}^{<s-1>} - \gamma \Delta \mathbf{W}_{\theta.t}^{\prime},
	\label{eq:update}
\end{equation}
where \({}^{<s>}\) denotes the \(s\)-th optimization step and \( \gamma \) is the learning rate.
The same procedure can be similarly extended to all layers of the ViT architecture.
Notably, this adaptation is optimizer-agnostic and can theoretically be integrated with any optimization algorithm. In our experiments, we employ the widely-used Adam optimizer, which empirically demonstrates satisfactory performance.

To implement \eqname~\eqref{eq:grad_mod}, our method employs the mask \( \bar{\mathbf{M}}_{t-1} \) constructed from the attention map \( \mathbf{U}_{t-1} \) of the previous task \( \mathcal{T}_{t-1} \).
The attention map should accurately capture the discriminative regions of the target object to ensure effective gradient modulation.
To this end, we develop a tailored procedure for attention map extraction and mask generation, as detailed in the following subsection.

\begin{table*}[t]
	\centering
	\setlength{\tabcolsep}{2pt}
	\resizebox{\textwidth}{!}{%
		\begin{tabular}{@{}llcccccccc@{}}
			\toprule
			\multirow{2}{*}{Method} & \multirow{2}{*}{Venue} & \multicolumn{2}{c}{10S-ImageNet-R}             & \multicolumn{2}{c}{20S-ImageNet-R}             & \multicolumn{2}{c}{10S-CIFAR-100}              & \multicolumn{2}{c}{10S-DomainNet}                \\ \cmidrule(l){3-10} 
			&                        & \acc                     & \forg               & \acc                     & \forg               & \acc                     & \forg               & \acc                     & \forg                 \\ \midrule
			L2P                     & CVPR'22                & 61.57\pms{0.66}          & 9.73\pms{0.47}      & 59.38\pms{0.50}          & 5.89\pms{0.36}      & 83.83\pms{0.04}          & 7.63\pms{0.30}      & ~\,81.17\pms{0.83}\tdr   & ~~8.98\pms{1.25}      \\
			DualPrompt              & ECCV'22                & 68.13\pms{0.49}          & 4.68\pms{0.20}      & 63.21\pms{0.49}          & 5.28\pms{0.45}      & 86.51\pms{0.33}          & 5.16\pms{0.09}      & ~\,81.70\pms{0.78}\tdr   & ~~8.04\pms{0.31}      \\
			CODA-Prompt             & CVPR'23                & 75.45\pms{0.56}          & 1.64\pms{0.10}      & 72.37\pms{1.19}          & 0.96\pms{0.15}      & 86.25\pms{0.74}          & 1.67\pms{0.26}      & ~\,80.04\pms{0.79}\tdr   & 10.16\pms{0.35}       \\
			APG                     & ICCV’23                & 73.27\quad~~~\,          & 8.59\quad~~~\,      & 71.22\quad~~~\,          & 7.39\quad~~~\,      & 89.35\quad~~~\,          & 6.01\quad~~~\,      & -                        & -                     \\
			ESN                     & AAAI'23                & 71.07\pms{0.29}          & 4.99\pms{0.49}      & 64.77\pms{0.71}          & 6.65\pms{1.24}      & 86.34\pms{0.52}          & 4.76\pms{0.14}      & ~\,79.22\pms{2.04}\tdr   & 10.62\pms{2.12}       \\
			OS-Prompt++             & ECCV'24                & 75.67\pms{0.40}          & 1.27 \pms{0.10}     & 73.77\pms{0.19}          & 0.79\pms{0.07}      & 86.68\pms{0.67}          & 1.18\pms{0.21}      & -                        & -                     \\
			EvoPrompt               & AAAI'24                & 76.83\pms{0.08}          & 2.78\pms{0.06}      & 74.41\pms{0.23}          & 2.56\pms{0.22}      & 87.97\pms{0.30}          & 2.60\pms{0.42}      & ~\,79.50\pms{0.29}\ddr   & ~~3.81\pms{0.36}      \\
			PGP                     & ICLR'24                & 69.34\pms{0.05}          & 4.53\pms{0.04}      & -                        & -                   & 86.92\pms{0.05}          & 5.35\pms{0.19}      & ~\,80.41\pms{0.25}\ddr   & ~~8.39\pms{0.18}      \\
			OVOR-Deep               & ICLR'24                & 76.11\pms{0.21}          & 7.16\pms{0.34}      & 73.85\pms{0.29}          & 6.80\pms{0.65}      & 85.99\pms{0.89}          & 6.42\pms{2.03}      & ~\,79.61\pms{0.86}\ddr   & ~~4.77\pms{0.94}      \\
			ConvPrompt              & CVPR'24                & 77.86\pms{0.25} 			& 4.33\pms{0.24}      & 75.10\pms{0.39} 		 & 4.10\pms{0.29}      & 88.87\pms{0.33} 		  & 4.75\pms{0.15}      & ~\,79.47\pms{0.35}\ddr   & ~~6.49\pms{0.43}      \\
			InfLoRA                 & CVPR'24                & 75.65\pms{0.14}          & 5.73\pms{0.44}      & 71.01\pms{0.45}          & 6.83\pms{0.44}      & 87.06\pms{0.25}          & 6.22\pms{0.39}      & ~\,81.45\pms{0.68}\ddr   & ~~5.35\pms{0.52}      \\
			EASE                    & CVPR'24                & 76.17\tsb{\qquad~~~}     & 7.82\tsb{\qquad~~~} & 73.27\tsb{\qquad~~~}     & 8.51\tsb{\qquad~~~} & 87.76\tsb{\qquad~~~}     & 5.94\tsb{\qquad~~~} & 78.89\ddr\tsb{\qquad~}   & ~~7.89\tsb{\qquad~~~} \\
			CPrompt                 & CVPR'24                & 77.14\pms{0.11}          & 5.97\pms{0.68}      & 74.79\pms{0.28}          & 7.34\pms{0.65}      & 87.82\pms{0.21}          & 5.06\pms{0.50}      & 82.97\pms{0.34} 		   & ~~7.45\pms{0.93}      \\
			VPT-CPG           		& AAAI'25                & \textud{78.63}\pms{0.52} & 7.18\pms{0.62}      & ~\,75.33\pms{0.49}\str   & 7.08\pms{0.52} 	   & \textud{90.63}\pms{0.44} & 3.98\pms{0.65}      & \textud{83.21}\pms{0.67} & ~~7.09\pms{0.82}      \\
			CAPrompt          		& AAAI'25                & ~\,71.67\pms{0.45}\str   & 5.24\pms{0.48}      & ~\,71.24\pms{0.80}\str   & 6.45\pms{0.78}      & ~\,87.27\pms{0.39}\str   & 4.90\pms{0.22}      & ~\,80.97\pms{0.98}\str   & ~~4.21\pms{0.34}        \\
			SD-LoRA           		& ICLR'25                & 77.34\pms{0.35}          & -				      & 75.26\pms{0.37}          & -			       & 88.01\pms{0.31}          & -      				& ~\,81.01\pms{0.42}\str   & -			           \\
			BiLoRA           		& CVPR'25                & 77.95\pms{0.14}          & -				      & 72.41\pms{0.76}          & -			       & 87.46\pms{0.76}          & -			        & 76.56\pms{0.41}          & -					   \\
			LoRA-DRS           		& CVPR'25                & 75.94\pms{0.46}          & -				      & 73.78\pms{0.44}          & -			       & 89.14\pms{0.23}          & -			        & ~\,71.19\pms{0.55}\str   & -        			   \\
			CPrompt-KAC        		& CVPR'25                & 78.07\quad~~~\,          & -				      & \textud{75.73}\quad~~~\, & -			       & 87.19\quad~~~\,		  & -			        & -   					   & -        			   \\
			SEMA	           		& CVPR'25                & 78.00\quad~~~\,          & -				      & 74.53\quad~~~\,		     & -      			   & 86.98\quad~~~\,		  & -			        & ~\,80.92\str\quad~~~\,   & -			           \\\midrule
			Seq-FT					& Baseline               & 49.43\pms{1.72}          & 44.31\pms{0.94}     & 39.63\pms{2.43}          & 53.98\pms{2.23}     & 52.85\pms{1.89}          & 49.24\pms{2.11}     & 44.06\pms{0.87}          & 59.43\pms{1.15}       \\
			ARCL-ViT              & This work              & \textbf{81.17}\pms{0.59} & ~~4.84\pms{1.45}    & \textbf{79.40}\pms{0.56} & ~~6.59\pms{1.01}    & \textbf{90.87}\pms{0.26} & ~~3.92\pms{0.34}    & \textbf{83.94}\pms{0.47} & ~~7.04\pms{1.21}      \\ \bottomrule
		\end{tabular}
	}
		\caption{Comparison with existing methods. The standard deviation (\ie, the value after \(\pm\)) is reported when available. Missing results in the corresponding papers are denoted by ``-''. Accuracies marked with \textdagger, \textdaggerdbl\, and * are reproduced by \cite{ConsistentPrompting2024Gao}, \cite{TrainingConsistent2025Lu} and us, respectively. `Seq-FT' represents the sequential fine-tuning baseline using \eqname~\eqref{eq:grad_nomod}. The best accuracy among all methods is highlighted in \textbf{bold}, and the best accuracy among existing methods is \textud{underlined}.}
		\label{tab:sota}
\end{table*}

\subsection{Adaptive Mask Generation}
We propose a layer-wise rollout approach based on the rollout method~\cite{QuantifyingAttention2020Abnara} to extract attention maps, as well as an adaptive mask construction strategy.
For the weight matrices \( \mathbf{W}_{q.t}^l, \mathbf{W}_{k.t}^l \) and \( \mathbf{W}_{v.t}^l \) in the \( l \)-th block (\( l \in \{1, 2, \ldots, L\} \)), the computation of its associated mask \( \bar{\mathbf{M}}_{t-1}^l \) is as follows.

We first consider a single sample from \( \mathcal{D}_{t-1} \) (omitting the sample identifier for brevity), which is fed into the model trained on \( \mathcal{T}_{t-1} \) for forward propagation. We extract the activated attention matrices from all layers, denoted as \( \{\mathbf{S}^1_{t-1}, \mathbf{S}^2_{t-1}, \ldots, \mathbf{S}^L_{t-1}\} \).
The layer-wise rollout attention matrix for the \( l \)-th layer is computed as:
\begin{equation}
	\hat{\mathbf{S}}_{t-1}^l = \tilde{\mathbf{S}}_{t-1}^{1} \cdot \tilde{\mathbf{S}}_{t-1}^{2} \cdot \ldots \cdot \tilde{\mathbf{S}}_{t-1}^{l},
	\label{eq:our_rollout}
\end{equation}
where \( \tilde{\mathbf{S}}_{t-1}^{l} = \mathbf{I} + \mathbf{S}_{t-1}^l \), and \( \mathbf{I} \) is the identity matrix to prevent self-suppression across layers~\cite{QuantifyingAttention2020Abnara}.
Based on \( \hat{\mathbf{S}}_{t-1}^{l} \!\in\! \mathbb{R}^{(N+1)^2} \), we extract its class attention map \( \mathbf{U}_{t-1}^{l} \!\in\! \mathbb{R}^{\sqrt{N}^2} \) by selecting the attention weights between the class token (query) and all image tokens (keys) and reshaping it into a matrix.
Specifically, \( \mathbf{U}_{t-1}^{l} \) corresponds to the first row of \( \hat{\mathbf{S}}_{t-1}^{l} \) excluding the first element, and can be formulated as \( \mathbf{U}_{t-1}^{l} \!=\! \hat{\mathbf{S}}_{t-1}^{l}[1, 2\!:] \).
We then construct the mask based on \( \mathbf{U}_{t-1}^{l} \).

Since the size of discriminative regions varies across images, using a fixed threshold or proportion for mask generation is suboptimal.
To address this, we propose an adaptive approach to distinguish between attention and background regions.
Due to the sharpening effect of the softmax function, the activated values in the attention regions are typically much higher than those in the background.
As a result, sorting the elements of \( \mathbf{U}_{t-1}^{l} \) in ascending order yields a curve that resembles a sigmoid function.
The optimal threshold index \( k^* \) is determined by locating the point with the lowest second-order derivative in the sorted attention curve:
\begin{equation}
	k^* = \mathop{\arg\min}\limits_{2 \leq k \leq N-1} \left( u_{k+1} - 2 u_k + u_{k-1} \right),
	\label{eq:adaptive_thres}
\end{equation}
where \( u_1, u_2, \ldots, u_{N} \) denotes the sorted elements of \( \mathbf{U}_{t-1}^{l} \), and the corresponding adaptive threshold is \( \tau_{t-1}^l = u_{k^*} \).
This point corresponds to the inflection point where the curve rises sharply from low values to the high-value plateau region.
Next, all elements in \( \mathbf{U}_{t-1}^{l} \) greater than or equal to \( \tau_{t-1}^l \) are classified as attention regions, and a binary mask \( \mathbf{M}_{t-1}^l \) is generated accordingly:
\begin{equation}
	\mathbf{M}_{t-1, i}^l =
	\begin{cases}
		0, & \text{if } \mathbf{U}_{t-1, i}^l \geq \tau_{t-1}^l \\
		1, & \text{otherwise}
	\end{cases}
	~\forall i \in \{1, \cdots, N\}
	\label{eq:mask}
\end{equation}

To apply \( \mathbf{M}_{t-1}^l \!\in\! \mathbb{R}^{\sqrt{N}^2} \) to \( \nabla(\mathbf{A}_t^l) \!\in\! \mathbb{R}^{(N+1)^2} \) for gradient masking, we extend and align the mask as follows.
First, we flatten \( \mathbf{M}_{t-1}^l \) into a row vector denoted as \( \mathbf{m}_{t-1}^l \), and prepend a 0 to its left which aims to maintain the stability of the class token, resulting in \( \bar{\mathbf{m}}_{t-1}^l \!\in\! \mathbb{R}^{N+1} \).
Next, to enforce a consistent attention retention rule for all queries, we broadcast \( \bar{\mathbf{m}}_{t-1}^l \) to all \( N+1 \) rows, yielding the mask matrix \( \bar{\mathbf{M}}_{t-1}^l \) of shape \( (N+1)^{2} \).
This mask is used in \eqname~\eqref{eq:grad_mod} for element-wise multiplication with \( \nabla(\mathbf{A}_t^l) \) and \( \mathbf{S}_t^l \) during gradient masking.

In practice, we further extend \( \mathbf{M}_{t-1}^l \) from each instance to all samples of a class and previous tasks by averaging, resulting in continuous masks within \([0,1]\).
For Transformers with multi-head attention, \eqname~\eqref{eq:grad_mod}\(\sim\)\eqref{eq:mask} can be applied independently to each head.
More details and the full algorithm are provided in the Appendix.

\section{Experiments}
\subsection{Experimental Setups}
We evaluate our method on four class-incremental learning (CIL) benchmarks, including 10-split and 20-split ImageNet-R, 10-split CIFAR-100 and 10-split DomainNet.
Note that the 10-split DomainNet benchmark was organized by \cite{IsolationImpartial2023Wang} for challenging cross-domain CIL, consisting of the 200 classes with the most images from the original DomainNet~\cite{MomentMatching2019Peng}.

We use ViT-B/16 pre-trained on ImageNet-21k~\cite{ImageNetLarge2015Russakovsky} as the backbone by default in our experiments.
Each task employs an independent classifier.
During training, the \( \mathbf{W}_{\{q,k,v\}} \) layers in all 12 ViT blocks as well as the classifier are fine-tuned.
We employ the Adam optimizer with a learning rate of 0.0001 for the \( \mathbf{W}_{\{q,k,v\}} \) layers and 0.01 for the classifier.
During inference, all classifiers are concatenated, and the backbone from the final task is used for prediction.
We report the final average accuracy and final average forgetting, each computed over three runs with different random seeds.
Additional experimental details are provided in the Appendix.
Our code is available at \url{https://github.com/zugexiaodui/AttentionRetentionCL}.
 
\begin{table}[t]
	\centering
	\setlength{\tabcolsep}{1.2pt}
	\begin{tabular}{@{}c|ccc|cc|cccc@{}}
		\toprule
		\multirow{2}{*}{Index} & \multicolumn{3}{c|}{~Attention} & \multicolumn{2}{c|}{~Mask} & \multicolumn{2}{c}{10S-ImageNet-R} & \multicolumn{2}{c}{10S-DomainNet} \\ \cmidrule(l){2-10} 
		& \textit{a.}    & \textit{b.}   & \textit{c.}   & \textit{d.}              & \textit{e.}           & Acc.               & Forg.    & Acc.              & Forg.    \\ \midrule
		1                      & \notuse          & \notuse         & \notuse              & \notuse            & \notuse            & 49.43              & 44.31         & 44.06             & 59.43         \\
		2                      & \use             & \notuse         & \notuse              & \use               & \notuse            & 77.63              & 12.84         & 76.28             & 20.86         \\
		3                      & \use             & \notuse         & \notuse              & \notuse            & \use               & 80.33              & ~~6.31        & 82.17             & 10.33         \\
		4                      & \notuse          & \use            & \notuse              & \notuse            & \use               & 80.43              & ~~7.42        & 82.28             & 11.10         \\
		5                      & \notuse          & \notuse         & \use                 & \use               & \notuse            & 76.72              & 13.86         & 75.60             & 21.61         \\
		6                      & \notuse          & \notuse         & \use                 & \notuse            & \use               & \textbf{81.17}     & ~~4.84        & \textbf{83.94}    & ~~7.04        \\ \bottomrule
	\end{tabular}
	\caption{Ablation study on different attention map extraction (\textit{a.}\,raw attention, \textit{b.}\,naive rollout, \textit{c.}\,layer-wise rollout) and mask thresholding (\textit{d.}\,fixed, \textit{e.}\,adaptive) strategies.}
	\label{tab:ablation_attn}
\end{table}

\begin{table}[t]
	\centering
	\setlength{\tabcolsep}{5pt}
	\begin{tabular}{@{}ccccc@{}}
		\toprule
		\multirow{2}{*}{Masked regions} 	& \multicolumn{2}{c}{10S-ImageNet-R} & \multicolumn{2}{c}{10S-DomainNet} \\ \cmidrule(l){2-5} 
		& Acc.               & Forgetting    & Acc.              & Forgetting    \\ \midrule
		Random                                 	& 73.27              & 4.22          & 79.52             & 6.70          \\
		Non-attention                          	& 69.55              & 7.56          & 76.08             & 9.47          \\
		Attention                              	& \textbf{81.17}     & 4.84          & \textbf{83.94}    & 7.04          \\ \bottomrule
	\end{tabular}
	\caption{Ablation study on different masked regions for attention maps.}
	\label{tab:ablation_zero}
\end{table}

\subsection{Comparison with Existing Methods}
We compare the proposed ARCL-ViT with existing methods on the four benchmarks in \tablename~\ref{tab:sota}.
The competitors include recent state-of-the-art works such as VPT-CPG~\cite{TrainingConsistent2025Lu}, SD-LoRA~\cite{SDLoRAScalable2025Wu}, BiLoRA~\cite{BiLoRAAlmostOrthogonal2025Zhu} and KAC~\cite{KACKolmogorovArnold2025Hu}.
The Seq-FT baseline refers to sequentially fine-tuning the \( \mathbf{W}_{\{q,k,v\}} \) layers and the classifier without any anti-forgetting technique.
The only difference between Seq-FT and our method is that Seq-FT updates parameters using \eqname~\eqref{eq:grad_nomod} while our method applies \eqname~\eqref{eq:grad_mod}\( \sim \)\eqref{eq:update}.
The direct comparison to Seq-FT validates the core effectiveness of our approach, which achieves 37\% higher accuracy and 46\% less forgetting by removing attention drift from previous tasks.
Furthermore, our method surpasses 15 recent approaches from 2024 and 2025 and achieves state-of-the-art performance, with up to 3.7\% and an average of 1.8\% higher accuracy across the benchmarks.
\begin{table}[t]
	\centering
	\setlength{\tabcolsep}{1.6pt}
	\begin{tabular}{@{}clcccc@{}}
		\toprule
		\multirow{2}{*}{Weights} & \multicolumn{1}{c}{\multirow{2}{*}{Method}} & \multicolumn{2}{c}{10S-ImageNet-R} & \multicolumn{2}{c}{10S-DomainNet} \\ \cmidrule(l){3-6} 
										& \multicolumn{1}{c}{}                        & Acc.               & Forgetting    & Acc.              & Forgetting    \\ \midrule
		\multirow{7}{*}{DINO-1k}     	& L2P                                         & 56.71              & 4.60          & 50.72             & 16.89         \\
										& CODA-P                                      & 64.02              & 7.70          & 66.74             & 8.91          \\
										& EASE                                        & 65.97              & 9.24          & 70.23             & 9.64          \\
										& CPrompt                                     & 64.66              & 9.26          & 75.31             & 9.93          \\
										& InfLoRA                                     & 68.31     		   & 8.81          & 75.05		       & 7.70          \\
										& VPT-CPG                                     & \textud{68.32}     & 2.78          & \textud{77.46}    & 10.18         \\ \cmidrule(l){2-6} 
										& Seq-FT                                      & 39.93              & 49.27         & 35.67             & 67.98         \\
										& ARCL-ViT                                  & \textbf{72.37}     & 4.91          & \textbf{79.39}    & 7.90          \\ \midrule
		\multirow{7}{*}{iBOT-1k}     	& L2P                                         & 60.80              & 6.65          & 53.32             & 17.12         \\
										& CODA-P                                      & 68.02              & 6.23          & 65.80             & 12.35         \\
										& EASE                                        & 69.91              & 9.22          & 72.39             & 10.25         \\
										& CPrompt                                     & 69.14              & 8.82          & 77.45             & 9.28          \\
										& InfLoRA                                     & 71.84     		   & 9.05          & 76.60    		   & 8.51          \\
										& VPT-CPG                                     & \textud{72.16}     & 5.80          & \textud{77.79}    & 13.02         \\ \cmidrule(l){2-6} 
										& Seq-FT                                      & 44.48              & 47.96         & 37.85             & 66.06         \\
										& ARCL-ViT                                  & \textbf{76.48}     & 7.68          & \textbf{80.88}    & 9.85          \\ \bottomrule
	\end{tabular}
	\caption{Results of using DINO and iBOT weights pre-trained on ImageNet-1k (\textit{i.e.}, DINO-1k and iBOT-1k).}
	\label{tab:pretraining}
\end{table}

\begin{table*}[t]
	\centering
	\setlength{\tabcolsep}{5pt}
	\begin{tabular}{@{}llcccccccc@{}}
		\toprule
		\multicolumn{1}{c}{\multirow{2}{*}{Method}} & \multicolumn{1}{c}{\multirow{2}{*}{Venue}} & \multicolumn{2}{c}{50S-ImageNet-R} & \multicolumn{2}{c}{50S-DomainNet} & \multicolumn{2}{c}{100S-ImageNet-R} & \multicolumn{2}{c}{100S-DomainNet} \\ \cmidrule(l){3-10} 
		\multicolumn{1}{c}{}                        & \multicolumn{1}{c}{}                       & Acc.                & Forgetting   & Acc.                & Forgetting  & Acc.                 & Forgetting   & Acc.                & Forgetting   \\ \midrule
		L2P                                         & CVPR'22                                    & 51.38               & 12.34        & 63.13               & 11.19       & 41.51                & 14.48        & 54.83               & 14.95        \\
		OVOR-Deep                                   & ICLR'24                                    & 63.25               & 5.23         & 68.29               & 6.85        & 43.02                & 7.30         & 52.09               & 8.66         \\
		ConvPrompt                                  & CVPR'24                                    & 64.61               & 7.12         & 71.76               & 6.37        & 44.32                & 8.97         & 56.21               & 6.27         \\
		InfLoRA                                     & CVPR'24                                    & 62.81               & 10.37        & 71.87			    & 9.20        & 42.23                & 14.17        & 48.06               & 15.43        \\
		EASE                                        & CVPR'24                                    & 70.27               & 6.73         & 65.34               & 9.64        & 51.56                & 7.56         & 37.26               & 29.12        \\
		CPrompt                                     & CVPR'24                                    & 70.75      		   & 7.44         & 70.74               & 9.01        & 59.90		         & 9.52         & 57.60   			  & 9.42         \\
		VPT-CPG                                     & AAAI'25                                    & \textud{73.08}      & 12.12        & \textud{72.27}      & 19.93       & \textud{64.63}       & 12.18        & \textud{60.81}      & 27.43        \\ \midrule
		Seq-FT                                      & Baseline                                   & 22.57               & 70.01        & 9.89                & 90.34       & 10.65                & 76.04        & 8.87                & 88.65        \\
		ARCL-ViT                                  & This work                                  & \textbf{76.45}      & 8.20         & \textbf{74.32}      & 15.95       & \textbf{70.72}       & 12.90        & \textbf{62.35}      & 25.63        \\ \bottomrule
	\end{tabular}
	\caption{Performance on long-sequence continual learning experiments using 50 and 100 splits of ImageNet-R and DomainNet.}
	\label{tab:long-seq}
\end{table*}

\subsection{Ablation Study}
\subsubsection{Construction of Attention Masks}
We investigate two factors in the construction of attention masks: 1) the method for extracting attention maps, and 2) the strategy for determining the mask threshold.
Specifically, three attention map extraction methods are considered:
\textit{a}) Raw attention in each layer, \textit{i.e.}, replacing \eqname~\eqref{eq:our_rollout} with \( \hat{\mathbf{S}}_{t-1}^l = \mathbf{S}_{t-1}^l \).
\textit{b}) Naive rollout, which simply accumulates the attention maps across all layers, \textit{i.e.}, \( \hat{\mathbf{S}}_{t-1}^l\!=\!\tilde{\mathbf{S}}_{t-1}^{1} \cdot \ldots \cdot \tilde{\mathbf{S}}_{t-1}^{L} \).
\textit{c}) Layer-wise rollout using \eqname~\eqref{eq:our_rollout}.
The thresholding strategy include two alternatives:
\textit{d}) Fixed threshold, where the top 20\% (determined by hyper-parameter search) positions in attention maps are set to 0.
\textit{e}) Adaptive threshold described in \eqname~\eqref{eq:adaptive_thres}.

The results of different configurations are presented in Table~2. Several key observations can be drawn:
1) Experiment 2, serving as a simplified variant of our approach, demonstrates the effectiveness of our core attention preservation concept for CL.
By employing basic raw-attention with fixed thresholding, this configuration achieves a substantial 30\% accuracy improvement and 35\% reduction in forgetting compared to the Seq-FT baseline (Index 1).
2) Layer-wise rollout attention exhibits superior compatibility with our framework.
Comparing experiments 3, 4, and 6, layer-wise rollout achieves an average accuracy improvement of 1.3\% over both raw attention and naive rollout methods.
3) Comparison between experiments 2 \textit{vs.} 3 and 5 \textit{vs.} 6 shows that the adaptive thresholding strategy yields an average accuracy gain of 5.4\% over the fixed threshold.
This result highlights the importance of adopting adaptive thresholding, which removes the need for manual tuning and enables more accurate identification of discriminative regions.
When layer-wise rollout is combined with adaptive thresholding, our method achieves the best performance.

\subsubsection{Different Masked Regions}
To validate the necessity of masking \emph{attention} regions specifically (\ie, \eqname~\eqref{eq:mask}), we conduct two comparative experiments:
1) Random masking. We randomly select positions to be set to zero in the mask, with the number of masked pixels matching that of the attention pixels identified by \eqname~\eqref{eq:mask}.
2) Non-attention Masking. We set non-attention regions to 0 by modifying the condition in \eqname~\eqref{eq:mask} to \( \mathbf{U}_{t-1, i}^l\!<\!\tau_{t-1}^l \).
As shown in \tablename~\ref{tab:ablation_zero}, masking the non-attention regions yields the poorest performance and leads to severe forgetting.
Interestingly, it still performs better than Seq-FT, which is because the non-attention regions for one class may overlap with the attention regions of another class, thereby indirectly providing partial retention to relevant attention areas.
Our proposed approach, which removes the attention regions of previous tasks, substantially outperforms the random masking baseline.
This systematic comparison confirms the necessity of masking attention regions to effectively reduce attention drift for CL.

\subsection{Generalization Experiments}
\subsubsection{Different Pre-training Weights}
To validate the generalizability of our method across different pre-training weights, we follow InfLoRA \cite{InfLoRAInterferenceFree2024Liang} and conduct experiments using ViT-B/16 initialized with self-supervised DINO~\cite{EmergingProperties2021Caron} and iBOT~\cite{ImageBERT2022Zhou} weights pre-trained on ImageNet-1k (denoted as DINO-1k and iBOT-1k, respectively).
For comparison, we also reproduce 6 representative methods when official results are unavailable.
As presented in \tablename~\ref{tab:pretraining}, ARCL-ViT exhibits significant improvements over the Seq-FT baseline, achieving an average accuracy gain of 37.8\% while reducing forgetting by 50.2\%.
It surpasses existing approaches by 2.9\% in accuracy with DINO-1k weights and by 3.7\% with iBOT-1k weights.
These results verify that our approach effectively generalizes to diverse pre-training weights.
Our method also generalizes well across different ViT architectures, with experimental results provided in the Appendix.

\subsubsection{Long-Sequence Continual Learning}
We further evaluate the generalizability of our method in long-sequence CL scenarios.
In this setting, the ImageNet-R and DomainNet datasets, each containing 200 classes, are partitioned into 50 and 100 tasks.
As shown in \tablename~\ref{tab:long-seq}, which compares our results with seven competitors (implementations from VPT-CPG~\cite{TrainingConsistent2025Lu}), ARCL-ViT outperforms the previous state-of-the-art approach VPT-CPG by 3.3\% in accuracy.
Although our method exhibits comparable forgetting, this result indicates that it achieves an effective trade-off between plasticity and stability.
This enables our model to continually acquire new knowledge through controlled forgetting.
Consequently, our approach attains superior overall accuracy in the challenging long-sequence CL scenarios.

\subsection{Analysis of Attention Retention}
We visualize the attention maps for samples from \( \mathcal{T}_1 \) of 10-split ImageNet-R and DoaminNet.
As shown in \figurename~\ref{fig1}(a), our method effectively preserves the original attention on key visual concepts.
We further make a quantitative comparison of attention drift across different methods, as illustrated in \figurename~\ref{fig1}(b).
The attention drift value at the \(t\)-th task (\( t \!>\! 2 \)) is calculated as follows:
For each sample in \( \mathcal{D}_1 \), attention maps are extracted using both the \( \mathcal{T}_1 \) and \( \mathcal{T}_t \) models.
The Frobenius norm of their difference is calculated and then normalized by the Frobenius norm of the \( \mathcal{T}_1 \) attention map to serve as the attention drift value.
Our method achieves the lowest attention drift at the end of training (6.0\%), which is significantly lower than that of Seq-FT (25.9\%) and other competitors.
These results provide quantitative evidence that our method retains attention to previously learned visual concepts, thereby mitigating catastrophic forgetting.

\section{Conclusion}
In this work, we introduce a novel attention-retaining framework to address the challenge of catastrophic forgetting in continual learning for Vision Transformers.
Concretely, we propose a gradient masking method to preserve selective attention to previously learned visual concepts.
Our method employs a layer-wise rollout and adaptive thresholding strategy for attention mask generation, and incorporates optimizer-compatible updates to ensure effective training.
Extensive experimental results on multiple continual learning scenarios demonstrate both the superiority and generalizability of our approach.
In future work, we will extend this framework to other Transformer-based architectures and explore the integration of additional biological principles.

\section*{Acknowledgments}
This work was supported in part by the National Natural Science Foundation of China (NSFC) under Grant 62576282, 62476223, 62376218; in part by the National Key Research and Development Program of China under Grant 2024YFF1306501; in part by Innovation Capability Support Program of Shaanxi (Program No. 2024ZC-KJXX-043); in part by Natural Science Basic Research Program of Shaanxi Province (2024JC-DXWT-07).

\bibliography{references}

\end{document}